# CityLearn: Standardizing Research in Multi-Agent Reinforcement Learning for Demand Response and Urban Energy Management

José R. Vázquez-Canteli
The University of Texas at Austin
jose.vazquezcanteli@utexas.edu

Sourav Dey
University of Colorado Boulder
sourav.Dey@colorado.edu

Gregor Henze
University of Colorado Boulder
gregor.henze@colorado.edu

Zoltan Nagy
The University of Texas at Austin
nagy@utexas.edu

## ABSTRACT

Rapid urbanization, increasing integration of distributed renewable energy resources, energy storage, and electric vehicles introduce new challenges for the power grid. In the US, buildings represent about 70% of the total electricity demand and demand response has the potential for reducing peaks of electricity by about 20%. Unlocking this potential requires control systems that operate on distributed systems, ideally data-driven and model-free. For this, reinforcement learning (RL) algorithms have gained increased interest in the past years. However, research in RL for demand response has been lacking the level of standardization that propelled the enormous progress in RL research in the computer science community. To remedy this, we created CityLearn, an OpenAI Gym Environment which allows researchers to implement, share, replicate, and compare their implementations of RL for demand response. Here, we discuss this environment and The CityLearn Challenge, a RL competition we organized to propel further progress in this field.

## KEYWORDS

Reinforcement Learning, Building Energy Control, Smart Buildings, Smart Grid

## 1 Introduction

Increasing urbanization, the integration of additional renewable energy generation resources, and the outlook of electric vehicles (EVs) introduce new challenges for the control of the power grid. The effects of these problems can already be observed in places like California, where rolling blackouts are becoming increasingly frequent, especially during the Summer, when consumers need electricity the most and demand is at its highest. Some of these problems can be tackled through additional capital investments to oversize the power grid at the transmission and distribution levels and create a buffer for supply and demand volatility. However, proper control techniques can reduce the need for such investments significantly. In the US, buildings represent about 70% of the total electricity demand, and it is estimated that demand response has the potential for reducing peaks of electricity demand by roughly 7% to 27% depending on the region [1].

In addition, as the energy storage capacity at the distribution level increases significantly (i.e. due to batteries and EVs), proper control and coordination strategies can help improve security of supply and reduce the cost of electricity by shaving the peak demand, aligning it with the peaks of renewable energy generation, and reducing the need for investments in expanding power capacity. Although there is no "silver bullet" to decarbonize the energy sector, energy flexibility and storage certainly play a role in the process [2]. The value of energy storage and demand response can be assessed as a function of multiple factors, including the energy mix, how volatile the loads are, the costs of different energy generation and storage technologies, and what control systems are leveraged. Similarly, there are multiple tradeoffs between using centralized or distributed energy resources, which depend on the incremental unit costs of distributed energy resources and the incremental locational value of electricity [3]. Thus, there is a need for more simulation tools that can help in such kinds of assessments under diverse sets of conditions.

Plenty of research has been conducted in the development of control methods for demand response [4], among which data-driven and model-free techniques, such as reinforcement learning, have gained popularity due to their potential cost-effectiveness and plug-and-play capabilities. However, more research is needed to improve the performance and reliability of RL algorithms for real-world implementation.

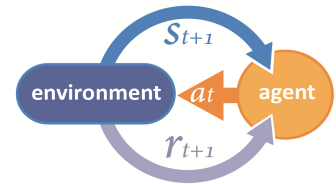

**Figure 1.** Agent-environment interaction in RL

Reinforcement learning (RL) is an agent-based and potentially model-free control algorithm that can learn by interacting with the environment under control [5] (figure 1). The goal of the agent (controller) is the maximization of the expected cumulative sum of the discounted rewards given an infinite time-horizon. RL problems are mathematically defined following a Markov decision process (MDP), which has sets of states $S$, actions $A$, a reward function $R: S \times A$ and transition probabilities between the states, $P: S \times A \times S \in [0,1]$. The policy $\pi$ maps the states and the actions, $\pi: S \rightarrow A$, and the value function $V^\pi(s)$ is the expected return for the agent if it starts in state $s$ and follows the policy $\pi$ thereafter, i.e.,

$$V^\pi(s) = \sum_a \pi(s,a) \sum_{s'} P^a_{s\,s'}[R^a_{s\,s'} + \gamma V^\pi(s')] \quad (1)$$

where $R^a_{s\,s'}$, is denoted as $r(s,a)$, and is the reward that the agent receives after taking an action $a = \pi(s_k)$, and transitioning from the current state $s$ to the next state $s'$, and $\gamma \in [0,1]$ is the discount factor for future rewards. An agent with $\gamma = 1$ will consider future rewards as important as current rewards, whereas for $\gamma = 0$, greater values are assigned to states leading to immediate rewards.

In model-free RL, the environment's dynamics (transition probabilities $P$) are unknown. Q-learning is likely the most known model-free RL algorithm [6]. For tasks that involve small state sets, the transitions can be expressed using a table that stores the state–action values (Q-values). Each entry in the table is a state–action tuple $(s, a)$, and the Q-values are updated as

$$Q(s_t, a_t) \leftarrow Q(s_t, a_t) + \alpha[r(s_t, a_t) + \gamma \max_a Q(s_{t+1}, a_t) - Q(s_t, a_t)] \quad (2)$$

where $s'$ is the next state, and $\alpha \in (0,1)$ is the learning rate, which defines to what extent new information overrides previous information and adjusts the learning speed. Q-values represent the expected cumulative sum of the discounted rewards after taking an action under a certain state and following a greedy policy thereafter. Q-learning is also an off-policy method, which can perform updates to its policy independently of the policy that it is currently following, by using past experience or historical data, which could have been collected using different policies. The collected state-action-reward tuples are stored in a replay buffer, from which they are sampled to perform the updates iteratively following (2).

Pure RL research has seen important developments in the last few years. This progress has been powered using standardized problems, or 'environments', that different research teams use to test and benchmark their RL agents and compare them with each other. This type of standardization has been missing in the research of RL for demand response and the control of urban energy systems. However, there are some instances of work that is being done towards the creation of testbeds for the implementation of RL in building energy systems. Moriyama et al. created an OpenAI Gym interface for EnergyPlus, which they have used as a testbed for the implementation of RL to control HVAC systems in datacenters [7]. Other work has focused on interfacing Modelica with an Open AI Gym module to create an environment for RL called Dymrl [8]. Similarly, ModelicaGym was created as an interface between Modelica and Open AI Gym environments [9].

These previous environments focus on the implementation of RL at the individual building level and can provide environments with passive energy storage control capabilities (i.e., control of the temperature setpoints, blinds, etc). However, for urban scale energy management and demand response, the control of active storage devices is often enough (i.e. thermal energy storage, batteries, EVs, etc.), which is the gap our research is addressing.

In this paper, we present and discuss CityLearn, an Open AI Gym environment for the implementation of RL for demand response and urban energy management [10]. The next Section describes the framework and building systems. In Section 3, we present application examples. One of them is The CityLearn Challenge 2020, an international RL challenge in which multiple research teams used RL to provide several simulated neighborhoods with load-shaping capabilities in four different climate zones. We also briefly introduce case studies that are already using CityLearn. Section 4 concludes the paper.

## 2 Methodology

### 2.1 Framework

CityLearn is a framework for the implementation of multi-agent or single-agent reinforcement learning algorithms for urban energy management, load-shaping, and demand response using the OpenAI Gym standard [11]. CityLearn is almost self-contained since it does not require co-simulation with EnergyPlus or any other building energy simulator. It only has a few dependencies (Python libraries such as Pandas must be installed). To achieve this, CityLearn uses building hourly data from pre-simulated models and assumes that the indoor temperatures of the building do not change as a function of the actions of the controllers. By not controlling passive storage (thermal mass of the building), the actions of the controller do not change the energy consumption of the building itself (excluding active storage), which allows building energy loads to be pre-simulated. Therefore, CityLearn guarantees that, at any time, the heating and cooling energy demand of the building are satisfied regardless of the actions of the controller. The actions of the reinforcement learning controller are automatically overridden to satisfy such constraints of thermal energy demand. This allows the controllers to focus on shaping the curve of electricity consumption without running the risk of interfering the comfort of the occupants or the desired temperatures.

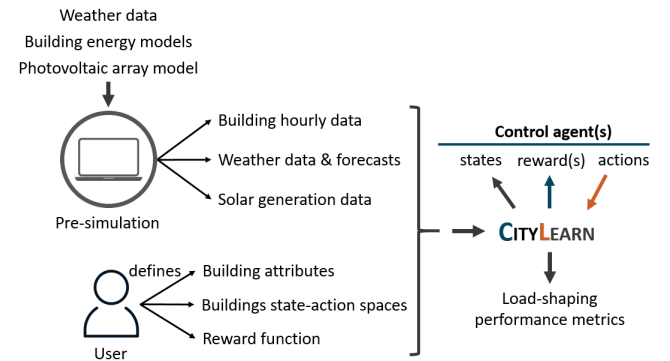

**Figure 2.** Flowchart of the CityLearn environment

As figure 2 shows, CityLearn reads input files that are obtained by pre-simulating the building energy models and the photovoltaic arrays under different weather conditions. It also takes as an input other files with various parameters defined by the user. These parameters include the attributes of the buildings and all the energy subsystem within them (heat pumps, batteries, thermal energy storage tanks, electric heaters, etc.), the states that each building will return every time-step, the actions that it accepts, and the reward function used by CityLearn to calculate and return the rewards of each building.

CityLearn's default mode is decentralized control, which is the mode to be used to implement multi-agent reinforcement learning controllers. Under this mode, CityLearn returns a list of rewards with as many rewards as buildings, and a list of lists with the states returned by each building. The actions must be inputted as a list of lists of actions.

Users can also select a centralized mode, in which the CityLearn environment will only return a single reward each time-step, and a list of states. The states returned are a sequence of unique states of all buildings. States such as outdoor temperature only appear once in the list of returned states, as it is the same for all buildings. Other state-variables such as the state-of-charge of the DHW tank, appear as many times as number of buildings there are. The actions are inputted in a list that contains all the actions for all the buildings. Therefore, under this mode, the dimensions of the state-action space increase linearly with the number of buildings.

All states, actions, and rewards are returned or inputted in the same order the buildings are sorted in the file "building_attributes.json". For more details about the state-variables available we refer the reader to the Appendix A.5.

## 2.2 CityLearn Overview

CityLearn is implemented in Python, and released open source on GitHub [12]. CityLearn has six important characteristics:

1. It allows both, single-agent, and multi-agent (decentralized) RL implementations.

2. The reward function is fully customizable and allows using a single-output function for single- output RL problems, and a multi-output function for multi-agent RL implementations.

3. Modular and open source: different energy systems' classes are provided and can be easily implemented. Users can also write their own classes and implement them easily, for example for district scale energy systems.

4. Easy to install and use: CityLearn requires very few dependencies, no co-simulation is needed, and the environment can easily be readily used by users from both the building/demand response domain and the computer science domain who are interested in solving problems with RL or multi-agent RL.

5. Users can create their own datasets (i.e. weather data, building energy demand, EV schedules) and feed them into CityLearn. Further instructions on how to create your own datasets can be found in the GitHub repository.

6. Speed: The energy loads and schedules are pre-computed (i.e. using EnergyPlus, Modelica, real-world data, etc) and provided to CityLearn as a dataset. Since CityLearn does not allow zone-level control (e.g., modifying temperature or relative humidity setpoints), the energy loads only need to be read, not calculated, during the simulation. Instead, the RL agents focus on controlling active energy storage systems such as water heaters, cooling thermal storage, batteries, or EV charging schedules at the urban level.

**Figure 3.** CityLearn's hierarchical architecture

Figure 3 shows how CityLearn inherits methods and attributes from the OpenAI Gym super class, and also contains further attributes, methods, and subclasses (energy systems). A detailed description of these components is provided in Appendix A. In the following we discuss the building and energy systems subclasses, and their implementation. Figure 4 illustrates these energy systems in CityLearn and how they relate to each other.

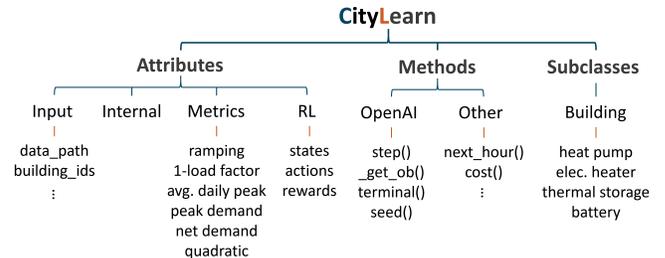

**Figure 4.** Energy models in CityLearn

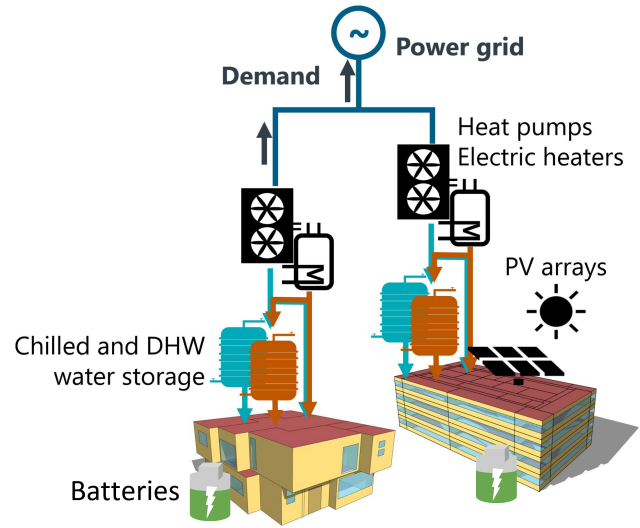

## 2.3 Building

As mentioned, every building is pre-simulated on an hourly timescale. Each building has its own non-RL control loop or schedule adjusting the temperature setpoints and other control variables within the HVAC system of the building. From the building energy simulations, the following variables are exported to be used by CityLearn:

- *month* – Month of the year (1 – 12)
- *hour* – Hour of day (1 – 24)

- *day_type* – Following EnergyPlus day type variable (1 = Sunday, 2 = Monday, …, 7 = Saturday, 8 = Holiday).
- *daylight savings status* – (0, 1)
- *indoor_average_temperature* – Average indoor temperature of the building across all thermal zones weighted by their floor area (°C).
- *average_unmet_cooling_setpoint_difference* – The unmet cooling setpoint difference (UCSD) is defined as the difference between the thermal zone temperature and its setpoint. Average UCSD is the average of the building across all the thermal zones and weighted by their floor area.
- *indoor_relative_humidity* – Average indoor relative humidity of the building across all thermal zones weighted by their floor area (0 - 100%).
- *equipment_electric_power* – Electricity consumed by all electrical appliances excluding the HVAC equipment (kWh).
- *heating_energy_for_DHW* – Thermal heating energy consumed for domestic hot water (kWh).
- *total_cooling_load* – Total thermal energy demand for cooling in the building (kWh).

At each time step every building is supplied with thermal energy from the thermal storage tanks and/or an energy supply unit such as a heat pump or an electric heater. To satisfy the assumption that the building temperature set-points are always met, which allows us to pre-simulate the buildings, the energy supply devices are sized to satisfy the energy demand of the building at any given time during the simulation. At the start of the simulation, the function *building_loader()* reads the output files from the pre-simulated building energy models and uses their maximum hourly thermal energy consumption to size the energy supply devices.

*Actions constraints*

Buildings also have a backup controller, different than the controller implemented by the user, that ensures that the building is always supplied with the thermal energy it needs to supply its temperature setpoints and DHW demand. This controller overrides the actions taken by the user-implemented controller if necessary: The user-implemented controller sends a control signal of how much energy it wants to charge or discharge from the thermal energy storage devices (chilled water and DHW). If the storage device is going to be charged, the energy supply device will do so always prioritizing that the energy demand of the building is met first. If the storage device is going to be discharged, it can only do so by an amount of energy that is no greater than the energy demand of the building.

The state of charge (SOC) is calculated every time-step in CityLearn for both cooling energy storage and DHW energy storage as follows:

$$SOC_{t+1} = SOC_t + \max\{\min\{a_t \cdot C, Q_{t_{max}} - Q_b\}, -Q^{dem}\} \quad (3)$$

$C \geq 0$: max. energy storage capacity of the sorage device

$-\frac{1}{C} \leq a_t \leq \frac{1}{C}$: user − implemented controller action

$Q^{dem} \geq 0$: building energy demand for cooling or DHW

$Q_{t_{max}} \geq Q^{dem}$ : max. thermal power of the heating device

Note that $Q_{t_{max}}$ can change over time in some cases (i.e. in a heat pump), as the outdoor air temperature has an impact on the coefficient of performance (COP) and this changes the maximum possible thermal power output.

Also note that the action of the user-implemented controller is bounded between $-\frac{1}{C}$ and $\frac{1}{C}$ because the capacity of the storage unit, $C$, is defined as a multiple of the maximum thermal energy consumption by the building in any given hour. For instance, if $C_{cooling} = 3$ and the peak cooling energy consumption of the building during the simulation is 20 kWh, then the storage unit will have a total capacity of 60 kWh. Therefore, the thermal storage device will never release more than $\frac{1}{C} 60\ kWh$ into the building. This limit should be set by the user in its own RL controller (for more efficient exploration). Otherwise, any action beyond the $-\frac{1}{C}$ and $\frac{1}{C}$ limits will always be overridden by the backup controller.

Unlike thermal energy storage devices, electric batteries do not have any load-based limits of operation. Their only limits of operation are given by the battery model itself as describes in subsection 2.7.

### 2.4 Heat pump

The heat pump model currently implemented in CityLearn is an air-to-water heat pump. Its coefficient of performance for cooling and heating are computed as:

$$COP_c = \eta_{tech} \cdot \frac{T^c_{target}}{T_{outdoor\ air} - T^c_{target}} \quad (4)$$

$$COP_h = \eta_{tech} \cdot \frac{T^h_{target}}{T^h_{target} - T_{outdoor\ air}} \quad (5)$$

Where $T^c_{target}$ and $T^h_{target}$ are the cooling and heating target temperatures respectively, and $\eta_{tech}$ is a technical efficiency coefficient (typically between 0.2 and 0.3). $T^c_{target}$ is equal to the logarithmic mean of the temperature of the supply water of the storage device and the temperature of the water returning to the heat pump. In CityLearn, the target temperatures are assumed to be constant and defined by the user. The value of $T^c_{target}$ typically ranges between 7 ºC and 10 ºC, and $T^h_{target}$ is often around 50 ºC.

The amount of thermal energy that the heat pump provides is based on the SOC from equation 6, and the demand for thermal energy by the building $Q^{dem}$.

$$Q^{hp}_{t+1} = C \cdot (SOC_{t+1} - SOC_t) + Q^{dem} \quad (6)$$

The air-to-water heat pump takes electrical energy from the grid, $E^{hp}$, and supplies thermal heating of cooling energy, $Q^{hp}$, to the building and/or the storage device following the equation:

$$E_t^{hp} = \frac{Q_t^{hp}}{COP_t} \quad (7)$$

## 2.5 Electric heater

The electric heater provides heating energy, $Q_{heater}$, (i.e. for DHW) consuming electricity from the grid, $E_{heater}$, and following the equation:

$$E_t^{heater} = \frac{Q_t^{heater}}{\eta_{eh}} \quad (8)$$

Where $\eta_{eh}$ is the user-defined heater efficiency and is usually greater than 0.9.

$Q_t^{heater}$ is obtained analogous to $Q_{t+1}^{hp}$ in equation 6.

## 2.6 Thermal storage

Buildings can have two different types of thermal energy storage: domestic hot water (DHW), and chilled water tanks. Each of them receives thermal heating or cooling energy from the energy supply devices, i.e., heat pump, electric heater. Figure 5 shows the energy flows of the thermal energy models and the names of the variables that we included in the equations.

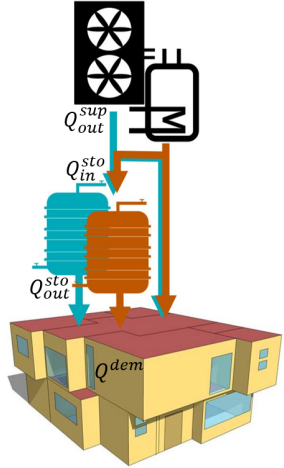

**Figure 5.** Thermal energy flows of CityLearn's energy models

By definition:

$$Q_{out}^{sup} \coloneqq Q_t^{hp}; \quad Q_{out}^{sup} \coloneqq Q_t^{heater}$$

The state of charge (SoC) is calculated following

$$SoC_{t+1} = SoC_t \cdot (1 - e_{loss}) + Q_{in}^{sto} - Q_{out}^{sto} \quad (9)$$

Where $e_{loss}$ is a thermal loss coefficient that indicates the fraction of thermal energy stored that is lost every hourly time-step. $\eta_{eff}$ is the round trip efficiency of the storage device.

$$Q_{out}^{sup} = \frac{Q_{in}^{sto}}{\sqrt{\eta_{eff}}} \quad (10)$$

$$Q_{out}^{sto} = -\frac{Q^{dem}}{\sqrt{\eta_{eff}}} \quad (11)$$

## 2.7 Battery

Unlike the thermal storage devices, the battery capacity is not defined as a multiple of the maximum hourly thermal energy demand of the building. Instead, the battery capacity, $C$, is defined in kWh. Its nominal power is defined in kW.

The battery model includes a capacity loss coefficient $c_{loss}$, which is the ratio of the capacity that gets lost due to the charge and discharge process, it is expressed in $\frac{1}{cycle}$ units. For example, $c_{loss} = 1e^{-5}$ would indicate that 0.001% of the battery's initial capacity is lost in each charge and discharge cycle. The new battery capacity, $C_{new}$, is given by the equation:

$$C_{new} = d \cdot \#\_of\_cycles \cdot C_0 \quad (12)$$

By replacing the variable $\#\_of\_cycles = \frac{|E_{in|out}|}{2 \cdot C}$ we obtain equation 13.

$$C_{new} = c_{loss} \cdot C_0 \cdot \frac{|E_{in|out}|}{2 \cdot C} \quad (13)$$

Where $C_0$ is the initial battery capacity, as defined by the user in kWh, $C$ is the current capacity, and $E_{in|out}$ is the amount of energy, in kWh, that has been charged or discharged.

The maximum charging power at any given time is given by the variable $P_t^{max}$, which depends on the state-of-charge SoC of the battery. The curve $P_t^{max}$ vs. SoC must be defined by the user in the battery "attribute capacity_power_curve" in the json file building_attributes. An example would be:

$$"capacity\_power\_curve": [[0., 1], [0.8, 1], [1., 0.2]]$$

In such example, the battery would charge at the nominal power rate $P_t^{max} = 1$ for state of the charge values $SoC \leq 0.8$ and would charge at 0.2 times the nominal power rate for values of the state of charge $0.8 < SoC \leq 1.0$. This is a reasonable example based on the assumption that some EVs (i.e. Tesla Model 3) can charge from 0% to 80% of their battery capacity in approximately 2 hours, but it takes them another 2 hours to charge from 80% to 100% at 220V [13].

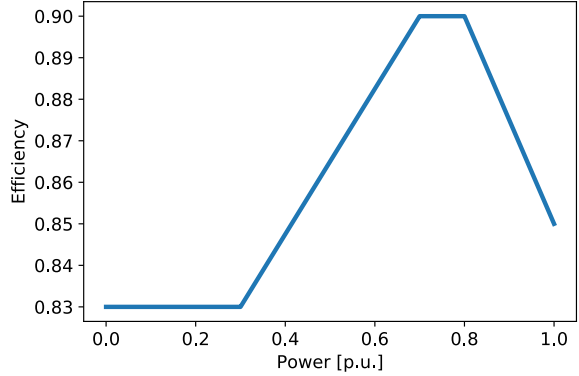

**Figure 6.** Efficiency vs. power curve. Where power is expressed on a per unit basis (p.u.) of the nominal power $P_t^{max}$ of the battery.

The round-trip efficiency $\eta_{eff}$ is also assigned by the user. It can be a constant value (battery attribute "efficiency") or a function of the charging and discharging rate P. This function can be defined in the battery attribute "power_efficiency_curve". An example would be:

$$"power\_efficiency\_curve": [[0.,0.83],[0.3,0.83],[0.7,0.9],$$
$$[0.8,0.9],[1.,0.85]]$$

In such example, the battery power vs. efficiency curve would look like in figure 6. This example is based on charging data provided by Tesla users [14], and it should not be used as a ground truth but just as an example on how to model these curves. The curve may also vary significantly depending on the model and technology of the battery.

The loss coefficient, $e_{loss}$ is another parameter that can be defined by the user and that represents the ratio at which the battery loses stored energy while being in stand-by. $e_{loss}$ is very low in batteries and it could be neglected ($e_{loss} \approx 0$).

$$SoC_{t+1} = SoC_t \cdot (1 - e_{loss}) + E_{in|out} \quad (14)$$

Where $E_{in|out} < 0$ if battery is being drawn from the battery.

If the RL agent takes an action that leads to a discharge of the battery by an amount that is greater than the electricity demand of the building, the excess electricity goes into the microgrid. Such excess electricity can be either used by other buildings (and reduce the amount of electricity drawn from the main feeder), or can go to the main grid (microgrid overgeneration). This behavior is described by equation 15 as follows:

$$E_{net}^{microgrid} = \sum_{i=0}^{n}(E_{b_i} + E_{bat}) \quad (15)$$

$$E_{bat} = \frac{E_{in|out_i}}{\sqrt{\eta_{eff}}} \quad if\ E_{in|out_i} \geq 0$$

$$E_{bat} = E_{in|out_i}\sqrt{\eta_{eff}} \quad if\ E_{in|out_i} < 0$$

Where n is the number of buildings in the microgrid, $E_{b_i}$ is the total net electricity consumption of the building, including the energy supply systems, and $E_{net}^{microgrid}$ is the net electricity consumption of the microgrid.

Electric vehicles (EVs), in our context, can be considered as an electric battery can therefore be modelled using the same equations above. In addition, a charging/discharging schedule must be defined.

### 2.8 Solar Photovoltaics

CityLearn uses pre-simulated data of photovoltaic generation per kW of installed solar PV power capacity. The user can then define the solar capacity installed in each building, in kW, and CityLearn multiplies this value by the pre-simulated data to obtain the final value of solar generation. The simulated datasets of photovoltaic currently available in our GitHub repository were obtained using SAM [15].

## 3 Application example

The GitHub repository of the CityLearn environment includes an application example of a multi-agent RL controller using the soft-actor critic (SAC) algorithm [16] to coordinate the energy consumption in a micro-grid of 9 buildings. The multi-agent algorithm is called MARLISA, and more details about how it allows buildings to coordinate with each other sharing limited information in a scalable, anonymous, and decentralized manner can be found in [17].

For this example, we used four datasets of 9 buildings each, consisting of the DOE prototype buildings [18]. The energy demand for each building has been pre-simulated using EnergyPlus in a different climatic zone of the USA (2A | Hot-Humid | New Orleans; 3A | Warm-Humid | Atlanta; 4A | Mixed-Humid | Nashville; 5A | Cold-Humid | Chicago) [19]. The group of buildings that we used is composed of one medium office (Building_1), one fast-food restaurant (Building_2), one standalone retail (Building_3), one strip mall retail (Building_4), and five medium multi-family buildings (Building_5 through Building_9).

Each building has an air-to-water heat pump and most buildings also have an electric heater that supplies them with DHW. All these devices, together with other electric equipment and appliances (non-shiftable loads) consume electricity from the main grid. Photovoltaic generation can offset part of this electricity consumption by allowing the buildings generate their own electricity.

When the building loads are pre-simulated it is important to use diverse load profiles to prevent the buildings from behaving very similar to each other. To create instances of the non-shiftable electrical loads of the residential buildings (Building_5 through Building_9), we used Pecan Street data [20] from multiple households in Austin, TX, and trained probabilistic regression models (neural networks with a soft-max output layer) and generated dozens of electricity profiles of appliances to feed the EnergyPlus models with. We used the same approach for the domestic hot water profiles using open-source data from the Solar Row project [21]. For more realistic energy consumption profiles, we generated diverse cooling and heating temperature setpoints for the different thermal zones of the multi-family buildings using data from the ResStock project [22].

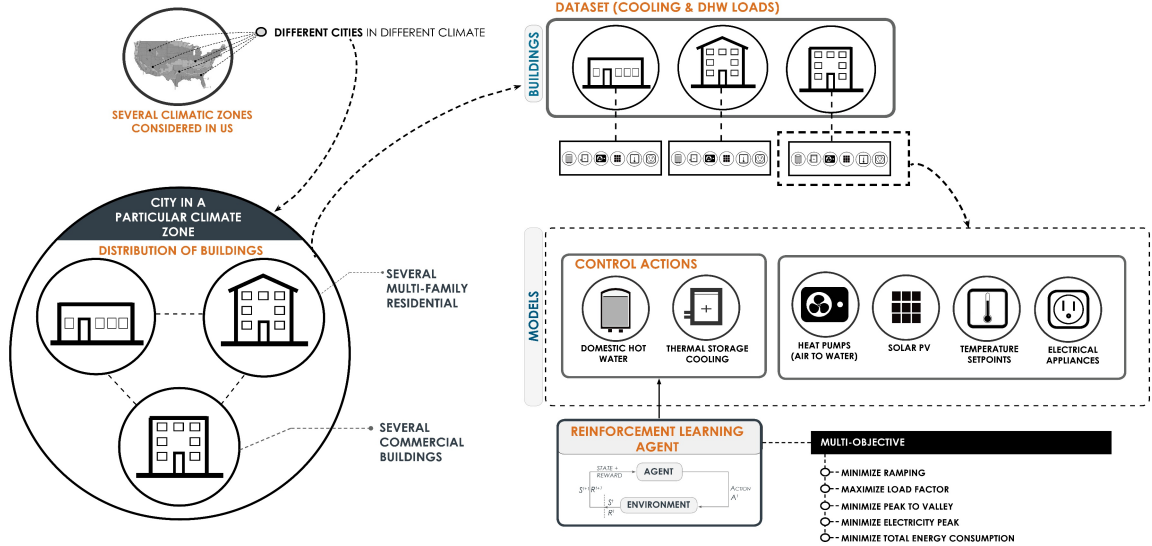

**Figure 7.** The CityLearn Challenge 2020 flowchart

### 3.1 The CityLearn Challenge

To promote our environment and test its potential for research standardization, and benchmarking of reinforcement learning algorithms, we launched The CityLearn Challenge 2020, an international challenge of reinforcement learning for load shaping and urban energy management. In this challenge, participants had to use reinforcement learning to control four microgrids, of 9 different buildings each, each microgrid located in a different climate zone. The objective of the agents was to minimize an average of all the load shaping metrics discussed in the Appendix A.8, except the quadratic metric. Figure 7 depicts the flowchart of the challenge.

During the challenge, a "design" dataset was made available to the participants, who used it to train their agents. Four teams then submitted their reinforcement learning agents to the organization team, which were run on a second, "evaluation" dataset. A leaderboard on the website of the challenge displayed the results of the different runs.

At the end of the challenge, all participants submitted their final agents, which were run on a third, "challenge" dataset to update the leaderboard with the final scores and find the winner, as illustrated in Table 1. All the scores are normalized by the scores of a benchmark rule-based controller (RBC). A score of 0.88 means that the RL controller performed 12% better than the baseline RBC. The datasets of the challenge are now public [23].

The version of the environment we provided for this challenge was like its current version but without electric batteries or the possibility of using the variable *net_electricity_consumption* as a state.

| Rank | Team | Climate Zone | Challenge Dataset (Final Scores) | | | | | |
|---|---|---|---|---|---|---|---|---|
| | | | Ramping | 1 - Load factor | Avg. Daily peak | Peak demand | Net elec. consumption | Average Score |
| 1 | pikapika | 1 | 0.661 | 0.921 | 0.902 | 0.909 | 0.977 | 0.874 |
| | | 2 | 0.706 | 0.946 | 0.864 | 0.998 | 0.979 | 0.899 |
| | | 3 | 0.679 | 0.870 | 0.830 | 0.857 | 0.978 | 0.843 |
| | | 4 | 0.695 | 0.910 | 0.892 | 1.015 | 0.992 | 0.901 |
| | | | | | | | Average Score | **0.879** |
| 2 | ProtoBuds | 1 | 0.779 | 1.014 | 0.982 | 1.131 | 1.015 | 0.984 |
| | | 2 | 0.780 | 0.980 | 0.959 | 0.999 | 1.013 | 0.946 |
| | | 3 | 0.812 | 0.960 | 0.939 | 1.083 | 1.018 | 0.962 |
| | | 4 | 0.860 | 0.996 | 0.991 | 1.013 | 1.017 | 0.976 |
| | | | | | | | Average Score | **0.967** |
| 3 | Learned City | 1 | 0.935 | 1.078 | 1.071 | 1.184 | 1.008 | 1.055 |
| | | 2 | 0.917 | 1.082 | 1.050 | 1.173 | 1.014 | 1.048 |
| | | 3 | 0.886 | 1.033 | 1.003 | 1.103 | 1.008 | 1.007 |
| | | 4 | 0.899 | 1.047 | 1.012 | 1.052 | 0.993 | 1.001 |
| | | | | | | | Average Score | **1.028** |
| 4 | EoRL | 1 | 0.929 | 1.131 | 1.150 | 1.220 | 1.044 | 1.095 |
| | | 2 | 0.940 | 1.095 | 1.105 | 1.254 | 1.046 | 1.088 |
| | | 3 | 0.898 | 1.064 | 1.053 | 1.221 | 1.049 | 1.057 |
| | | 4 | 0.892 | 1.065 | 1.069 | 1.107 | 1.035 | 1.034 |
| | | | | | | | Average Score | **1.068** |

**Table 1.** Final leaderboard of The CityLearn Challenge 2020

## 3.2 Case Studies

Since its inception [10], CityLearn has been used by multiple researchers to control building microgrids. The authors of [24] implemented a centralized soft-actor critic deep RL approach for demand response in a micro-grid using CityLearn. In [25], the author proposed and implemented in CityLearn an energy pricing agent with the objective of shifting peak electricity demand towards periods of peak renewable energy generation. In [26], the authors implemented four different RL control approaches in CityLearn and compare them among each other. Other researchers have focused on using other algorithms, inspired in RL, for demand response in microgrids. This is the case of [27], where the authors used CityLearn and compared their own control mechanism with the RL control approach.

## 4 Conclusion

In this paper we have introduced CityLearn, an OpenAI Gym environment, and a simulated framework for the implementation of reinforcement learning for demand response and urban energy management. While plenty of previous research in this field attempts provide energy savings or demand response capabilities without violating comfort constraints, CityLearn uses a different approach: by overriding the controllers' actions if necessary, it guarantees that the thermal energy demands of the buildings are always satisfied. This approach allows CityLearn to avoid any need for co-simulation since the buildings' energy demands can be pre-simulated and shared in CSV files among researchers, as we have done by publicly sharing the datasets of the challenge. This also makes it easier for researchers from the computer science domain to use this multi-agent reinforcement learning environment, as they only need to download it from the GitHub repository, and run their reinforcement learning agents as they would do with any other environment.

CityLearn also allows for customization, since users can select which buildings they want to control, which energy systems they have, and which states they return. Users can also design the reward function and choose between two different modes: central-agent and multi-agent mode. Regarding batteries and electric vehicles, CityLearn allows its users to customize the power vs. SoC, and power vs. efficiency curves among other characteristics. It also enables users to use pre-simulated data from complex building energy models, or even empirical data from actual buildings and load it in CityLearn. These modelling options improve the level of realism of CityLearn, which is an important guideline to follow when designing these kinds of environments [28].

Recently, machine learning experts looked for ways in which the machine learning community could help in tackling climate change, "one of the greatest challenges facing humanity" [29]. They identified the electrification of urban energy systems as a field where machine learning can provide a high leverage and help. CityLearn is aimed precisely at this objective, and a standardized environment will allow more researchers from the computer science and machine learning communities to participate in the search for solutions to tackle climate change.

# Appendix A

This appendix provides more details about the different concepts that are included in Figure 3.

## A.1 Input Attributes

- *data_path* - path indicating where the data is
- *building_attributes* - name of the file containing the characieristics of the energy supply and storage systems of the buildings
- *weather_file* - name of the file containing the weather variables
- *solar_profile* - name of the file containing the solar generation profile (generation per kW of installed power)
- *building_ids* - list with the building IDs of the buildings to be simulated
- *buildings_states_actions* - name of the file containing the states and actions to be returned or taken by the environment
- *simulation_period* - hourly time period to be simnulated. (0, 8759) by default: one year.
- *cost_function* - list with the cost functions to be minimized.
- *central_agent* - allows using CityLearn in central agent mode or in decentralized agents mode. If True, CityLearn returns a list of observations, a single reward, and takes a list of actions. If False, CityLearn will allow the easy implementation of decentralized RL agents by returning a list of lists (as many as the number of building) of states, a list of rewards (one reward for each building), and will take a list of lists of actions (one for every building).
- *verbose* - set to 0 if you don't want CityLearn to print out the cumulated reward of each episode and set it to 1 if you do

## A.2 Internal Attributes

- *net_electric_consumption* - district net electricity consumption
- *net_electric_consumption_no_storage* - district net electricity consumption if there were no cooling storage and DHW storage
- *net_electric_consumption_no_pv_no_storage* - district net electricity consumption if there were no cooling storage, DHW storage and PV generation
- *electric_consumption_dhw_storage* - electricity consumed in the district to increase DHW energy stored (when > 0) and electricity that the decrease in DHW energy stored saves from consuming in the district (when < 0).
- *electric_consumption_cooling_storage* - electricity consumed in the district to increase cooling energy stored (when > 0) and electricity that the decrease in cooling energy stored saves from consuming in the district (when < 0).
- *electric_consumption_dhw* - electricity consumed to satisfy the DHW demand of the district
- *electric_consumption_cooling* - electricity consumed to satisfy the cooling demand of the district
- *electric_consumption_appliances* - non-shiftable electricity consumed by appliances
- *electric_generation* - electricity generated in the district

## A.3 CityLearn Methods

- *get_state_action_spaces()* - returns state-action spaces for all the buildings
- *next_hour()* - advances simulation to the next time-step
- *get building information()* - returns attributes of the buildings that can be used by the RL agents (i.e. to implement building-specific RL agents based on their attributes, or control buildings with correlated demand profiles by the same agent)
- *get baseline cost()* - returns the costs of a Rule-based controller (RBC), which is used to divide the final cost by it.
- *cost()* - returns the normlized cost of the enviornment after it has been simulated. cost < 1 when the controller's performance is better than the RBC.

## A.4 Methods inherited from OpenAI Gym

- *step()* - advances simulation to the next time-step and takes an action based on the current state
- *_get_ob()* - returns all the states
- *_terminal()* - returns True if the simulation has ended
- *seed()* - specifies a random seed

## A.5 States

- *month* - 1 (January) through 12 (December)
- *day* - type of day as provided by EnergyPlus (from 1 to 8). 1 (Sunday), 2 (Monday), ..., 7 (Saturday), 8 (Holiday)
- *hour* - hour of day (from 1 to 24).
- *daylight_savings_status* - indicates if the building is under daylight savings period (0 to 1). 0 indicates that the building has not changed its electricity consumption profiles due to daylight savings, while 1 indicates the period in which the building may have been affected.
- *t_out* - outdoor temperature in Celcius degrees.
- *t_out_pred_6h* - outdoor temperature predicted 6h ahead (accuracy: +-0.3C)
- *t_out_pred_12h* - outdoor temperature predicted 12h ahead (accuracy: +-0.65C)
- *t_out_pred_24h* - outdoor temperature predicted 24h ahead (accuracy: +-1.35C)
- *rh_out* - outdoor relative humidity in %.
- *rh_out_pred_6h* - outdoor relative humidity predicted 6h ahead (accuracy: +-2.5%)
- *rh_out_pred_12h* - outdoor relative humidity predicted 12h ahead (accuracy: +-5%)
- *rh_out_pred_24h* - outdoor relative humidity predicted 24h ahead (accuracy: +-10%)
- *diffuse_solar_rad* - diffuse solar radiation in W/m^2.
- *diffuse_solar_rad_pred_6h*: diffuse solar radiation predicted 6h ahead (accuracy: +-2.5%)
- *diffuse_solar_rad_pred_12h* - diffuse solar radiation predicted 12h ahead (accuracy: +-5%)
- *diffuse_solar_rad_pred_24h* - diffuse solar radiation predicted 24h ahead (accuracy: +-10%)
- *direct_solar_rad* - direct solar radiation in W/m^2.

- *direct_solar_rad_pred_6h* - direct solar radiation predicted 6h ahead (accuracy: +-2.5%)
- *direct_solar_rad_pred_12h* - direct solar radiation predicted 12h ahead (accuracy: +-5%)
- *direct_solar_rad_pred_24h* - direct solar radiation predicted 24h ahead (accuracy: +-10%)
- *t_in* - indoor temperature in Celcius degrees.
- *avg_unmet_setpoint* - average difference between the indoor temperatures and the cooling temperature setpoints in the different zones of the building in Celcius degrees. sum((t_in - t_setpoint).clip(min=0) * zone_volumes)/total_volume
- *rh_in* - indoor relative humidity in %.
- *non shiftable load* - electricity currently consumed by electrical appliances in kWh.
- *solar_gen* - electricity currently being generated by photovoltaic panels in kWh.
- *cooling_storage_soc* - state of the charge (SOC) of the cooling storage device. From 0 (no energy stored) to 1 (at full capacity).
- *dhw_storage_soc* - state of the charge (SOC) of the domestic hot water (DHW) storage device. From 0 (no energy stored) to 1 (at full capacity).
- *net_electricity_consumption* - net electricity consumption of the building (including all energy systems) in the current time step.

### A.6 Actions

- *cooling_storage* - increase (action > 0) or decrease (action < 0) of the amount of cooling energy stored in the cooling storage device. -1.0 <= action <= 1.0 (attempts to decrease or increase the cooling energy stored in the storage device by an amount equal to the action times the storage device's maximum capacity). In order to decrease the energy stored in the device (action < 0), the energy must be released into the building's cooling system. Therefore, the state of charge will not decrease proportionally to the action taken if the demand for cooling of the building is lower than the action times the maximum capacity of the cooling storage device.
- *dhw_storage* - increase (action > 0) or decrease (action < 0) of the amount of DHW stored in the DHW storage device. -1.0 <= action <= 1.0 (attempts to decrease or increase the DHW stored in the storage device by an amount equivalent to action times its maximum capacity). In order to decrease the energy stored in the device, the energy must be released into the building. Therefore, the state of charge will not decrease proportionally to the action taken if the demand for DHW of the building is lower than the action times the maximum capacity of the DHW storage device.
- *battery_storage* - increase (action > 0) or decrease (action < 0) of the amount of electricity stored in the battery. -1.0 <= action <= 1.0 (attempts to decrease or increase the electricity stored in the battery by an amount equivalent to action times its maximum capacity). Each battery, when discharged, first prioritizes satisfying the electricity demand of the building where it belongs, and then uses any leftover discharged electricity to satisfy the electricity consumption at the microgrid level, causing overgeneration at the microgrid level is the energy discharged is greater than the rest of the net load of the microgrid.

### A.7 Rewards

For a central single-agent (if CityLearn class attribute central_agent = True):

- *reward_function_sa* – it takes the total net electricity consumption of each building (< 0 if generation is higher than demand) at every time-step as input and returns a single reward for the central agent.

For a decentralized multi-agent controller (if CityLearn class attribtue central_agent = False):

- *reward_function_ma* - class that can take building information and the number agents when instantiated. It contains a "get_rewards()" method that takes the total net electricity consumption of each building (< 0 if generation is higher than demand) at every time-step as input and returns a list with as many rewards as the number of agents.

*Important note: Users who want to implement multiple single-agent independent RL controllers (no communication among them nor collective rewards) must use central_agent = False, and define the class *reward_function_ma* in a way that the returned rewards for every agent are independent from each other.

### A.8 Evaluation metrics

env.cost() is the cost function of the environment, which the RL controller must minimize. There are multiple cost functions available, which are all defined as a function of the total non-negative net electricity consumption of the whole neighborhood:

- *ramping* - sum(|e(t)-e(t-1)|), where e is the net non-negative electricity consumption every time-step.
- *1-load_factor* - the load factor is the average net electricity load divided by the maximum electricity load.
- *average_daily_peak* - average daily peak net demand.
- *peak_demand* - maximum peak electricity demand
- *net electricity consumption* - total amount of electricity consumed
- *quadratic* - sum(e^2), where e is the net non-negative electricity consumption every time-step. (Not used in The CityLearn Challenge).